\documentclass{tADR2e}

\usepackage{url}
\usepackage{color}

\usepackage{natbib}

\begin{document}
\jvol{00} \jnum{00} \jyear{2019} \jmonth{January}

\articletype{Article}

\title{Team O2AS at the World Robot Summit 2018: \\
An Approach to Robotic Kitting and Assembly Tasks\\
using General Purpose Grippers and Tools}

\author{Felix von Drigalski$^{a}$$^{\ast}$, Chisato Nakashima$^{b}$, Yoshiya Shibata$^{b}$, Yoshinori Konishi$^{b}$\\
Joshua C. Triyonoputro$^{c}$, Kaidi Nie$^{c}$, Damien Petit$^{c}$, 
Toshio Ueshiba$^{d}$, Ryuichi Takase$^{d}$\\
Yukiyasu Domae$^{d}$, Taku Yoshioka$^{e,f}$, Yoshihisa Ijiri$^{a}$, 
Ixchel G. Ramirez-Alpizar$^{d,c}$, Weiwei Wan$^{c,d}$\\
and Kensuke Harada$^{c,d}$\thanks{$^\ast$Corresponding author: Felix von Drigalski, Email: f.drigalski@sinicx.com}\\\vspace{6pt}  
$^{a}${\em{OMRON SINIC X Corp., Tokyo, Japan}};\ \
$^{b}${\em{OMRON Corp., Kyoto, Japan}};\\
$^{c}${\em{Osaka University, Osaka, Japan}};\ \
$^{e}${\em{SenseTime Japan Ltd., Kyoto, Japan}};\\
$^{d}${\em{National Inst. of Advanced Industrial Science and Technology (AIST), Tokyo, Japan}}\\
$^{f}${\em{Laboro.AI Inc., Tokyo, Japan}}\\
\vspace{6pt}\received{v1.0 released January 2019} }

\maketitle

\begin{abstract}
In this article, we propose a versatile robotic system for kitting and assembly tasks which uses no jigs or commercial tool changers.
Instead of specialized end effectors, it uses its two-finger grippers to grasp and hold tools to perform subtasks such as screwing and suctioning.
A third gripper is used as a precision picking and centering tool, and uses in-built passive compliance to compensate for small position errors and uncertainty.
A novel grasp point detection for bin picking is described for the kitting task, using a single depth map.
Using the proposed system we competed in the Assembly Challenge of the Industrial Robotics Category of the World Robot Challenge at the World Robot Summit 2018, obtaining 4th place and the SICE award for lean design and versatile tool use.
We show the effectiveness of our approach through experiments performed during the competition.

\medskip

\end{abstract}

\section{Introduction} 
\label{sec:intro}

As products are becoming increasingly customized to reflect the tastes of individuals, robots are expected to manufacture larger and larger varieties of low-volume series of products.
The industrial robotics category of the World Robot Challenge (WRC), described in detail in~\cite{Yokokohji2019}, is a world-wide competition inciting teams to tackle increasing levels of automation in product manufacturing.
The competition aims at a quick and accurate assembly of model products with small tolerances, using highly autonomous and versatile robotic systems that need to respond to changing product specifications. 
In November of 2018, the WRC pre-competition was held at Tokyo Big Sight. 
OMRON Corp., OMRON SINIC X Corp., Osaka University, National Inst. of Advanced Industrial Science and Technology (AIST) and SenseTime Japan Ltd. participated in this competition together as team ``O2AS''.
In this paper, we explain our solution and our approaches for three of the four tasks of the competition:
\begin{itemize}
    \item The kitting task, where a robot needs to pick objects from bins placed on a table and arrange them in trays.
    \item The assembly task, where a robot is required to assemble a belt drive unit from the parts placed in trays.
    \item The assembly with surprise parts task, where a similar unit with different parts is revealed shortly before the competition, and has to be assembled in the same time as the original unit.
    \item The taskboard task, which recreates the elementary subtasks required for the assembly task (e.g. peg-in-hole, belt, screwing).
\end{itemize}
The strategies for the taskboard task, as well as the design of the precision gripper and several tools used to solve it, are explained in \cite{nie2019} in this journal issue.

\begin{figure}[h]
  \centering
  \includegraphics[width=140mm]{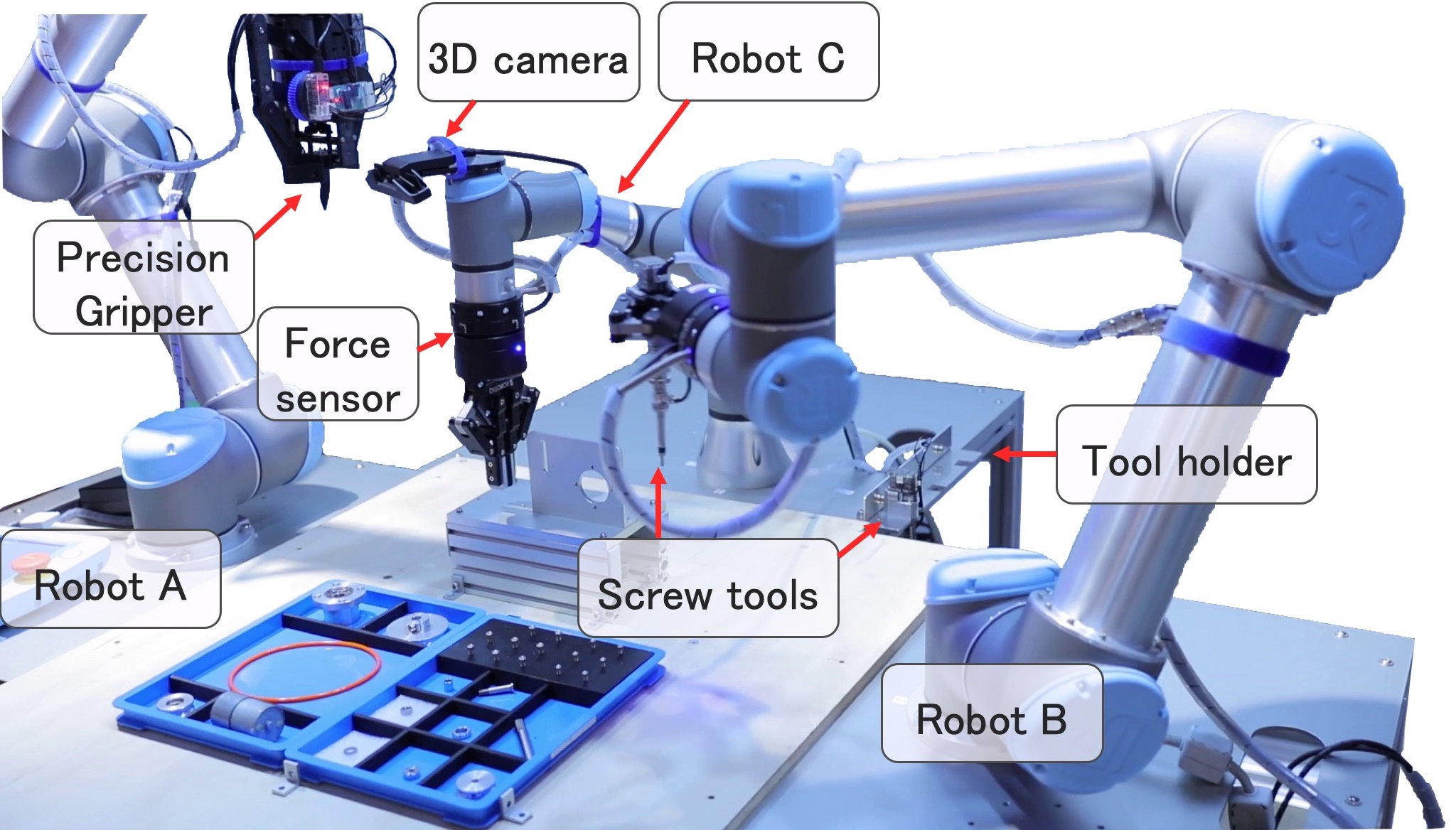}
  \caption{The robotic system used by team O2AS at the WRC performing the assembly task. Robot C is placing the plate while robot B is about to fasten it to the base plate using the screw tool and a screw that it picked from the blue parts trays in the foreground. A second screw tool is stored in the holder, which can be accessed by robots B and C. The three robots surround the assembly and parts trays, which they can manipulate simultaneously.}
  \label{fig:robot-system-photo}
\end{figure}

Flexible and automated product assembly has been an active research topic in robotics and AI for decades.
Lozano-Perez et al.~\cite{perez1987} first proposed the concept of Handy, where a robot performs the product assembly integrating the recognition, planning and manipulation capabilities.
Manipulation planners aiming at product assembly have been researched by many researchers such as~\cite{koga1994,perez2011,wan2016}. 
Robotic product assembly has also been in strong demand by the industry~\cite{IHI2009}.

However, in conventional robotic product assembly, the most common solution is to use grippers or gripper tips that match each part (e.g. a gripper tip with a concave opening with a curvature that matches the diameter of a cylinder to be grasped), and specially designed jigs that control the position and orientation of each part.
However, because these approaches require additional engineering to deal with new parts, their cost is prohibitive for small lot sizes and low-cost products, so they are only used in product lines producing many thousands of identical pieces.
We attempted to build a system that uses the same end effectors and tools to assemble a wide range of objects to overcome these limitations.

The most distinctive feature of our robotic system is its flexibility, which eliminates the usage of jigs and tool changers entirely.
Instead, we use general purpose grippers and custom tools that are picked up by the robot.
For example, our robots can grasp a screwing tool and use it to fasten a bolt.
To align a part for insertion, we use two robots to regrasp the part using only the grippers, without the use of jigs.
This makes the system highly flexible, and able to assemble a wide range of parts without any physical changes to the system.

The main contributions of this paper are the following:

\begin{itemize}
    \item A robot system which can kit and assemble small and reflective metal parts using no jigs or tool changers.
    \item Screw tools which can be grasped by general robotic grippers.
    \item An efficient grasping point detection algorithm for kitting that uses a single depth map.
    \item The source code running our solution~\footnote{Available at https://github.com/o2as/ur-o2as. Excluding proprietary vision modules.}.
\end{itemize}

The rest of this paper is organized as follows. 
In Section \ref{sec:system}, we explain the tools we have developed and the structure of our solution, including both hardware and software, as well as the reasons for our design choices.
In Section \ref{sec:approach}, we explain our approach to the kitting and assembly tasks, including the novel grasp point detection algorithm, and noting pitfalls that we have encountered and strategies that we found to be dead ends.
Data about our system's performance and experimental results is provided in Section \ref{sec:experiments}.
Finally, we discuss our system and the conclusions we drew from our experience in the competition in Section \ref{sec:discussion}.


\section{Robot System} 
\label{sec:system}

\begin{figure}
    \centering
    \includegraphics[width=\textwidth]{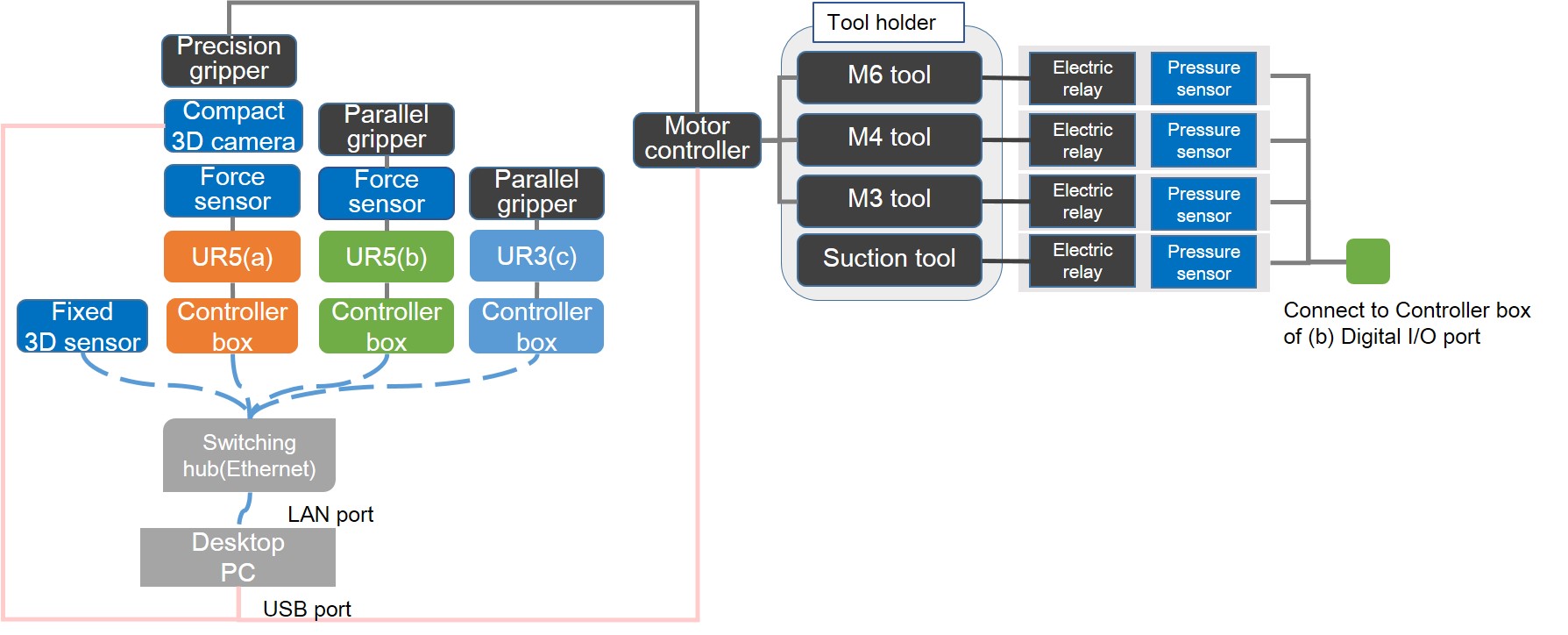}
    \caption{System diagram showing the wiring and hardware connections.}
    \label{fig:system_diagram}
\end{figure}

As shown in Figure~\ref{fig:robot-system-photo}, our robotic system is composed of one Universal Robots UR3 and two UR5 robot arms (6 DOF), and a set of 10 different tools.
The same robots and grippers were used in all three tasks only the tool arrangement in the holder was changed according to the task.
An overview of the system structure is shown in Figure~\ref{fig:system_diagram}.

We chose three arms rather than two to maintain flexibility during the design phase.
One of the robots was used to prototype gripper designs, and was not guaranteed to be able to grasp all sizes of parts and the tools.
As some operations such as fixing the screws to the base plate shown in Figure~\ref{fig:system_diagram} required one arm to hold the object while the other manipulates it, we opted to equip two robots with off-the-shelf grippers.
We also intended to use the third robot's camera for visual servoing to position items and tool tips, but discarded this idea later during the development due to time constraints.
In practice, we found that two robots would be sufficient, as long as they could grasp all the parts and tools.

\subsection{Robots and Grippers} 
\label{sec:system-hardware-robots-and-grippers}

The robot arms are mounted to the same structure, which consists of a base for each robot and a wooden board that connects all robot bases together and serves as a working surface.
The wooden board was chosen over a metal sheet so as to allow rapid prototyping and easy reconfiguration during the development phase, and is rigidly connected to the robot bases for easier calibration.

Robots B and C were equipped with off-the-shelf general purpose grippers~\footnote{Robotiq 2-Finger 85}, and robot A with a pincer-like precision gripper that uses compliance to pick, center and place small parts.
The general purpose grippers had a maximum holding force of 100~N and fingertips covered with 2~mm of rubber.

The precision gripper is passively compliant along the z-axis of the coordinate system (the axis pointing away from the robot wrist), which allowed the robot to plan to a location that is physically ``below'' the part (or inside the surface it rests on), thus ensuring contact with the surface and part.
Due to the gripper's design and the 3D printed parts, it was also compliant in the x- and y-axes, which was useful when performing spiral motions to compensate for positioning uncertainty.
The precision gripper design is detailed in \cite{nie2019} in this journal issue.

\begin{figure}[!t]
    \centering
    \includegraphics[height=100pt]{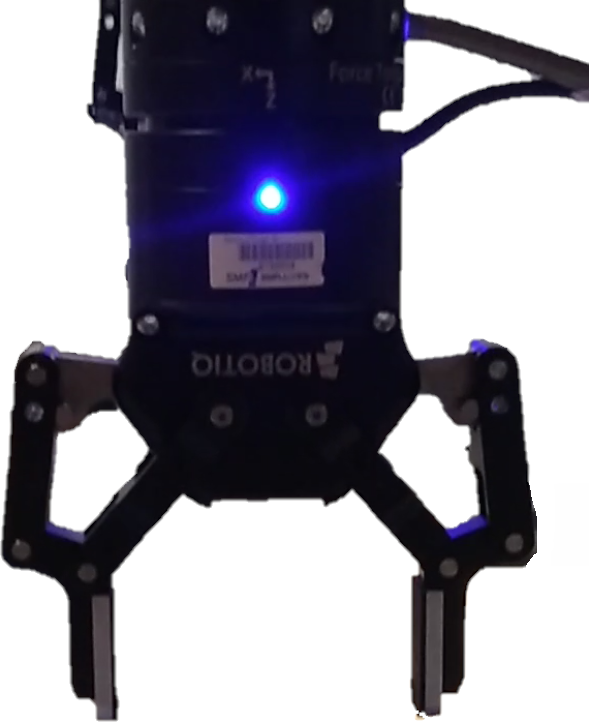}
    \includegraphics[height=100pt]{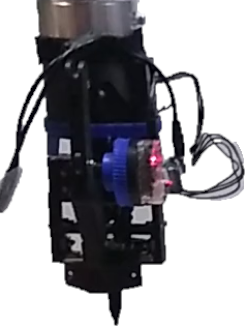}
    \includegraphics[height=100pt]{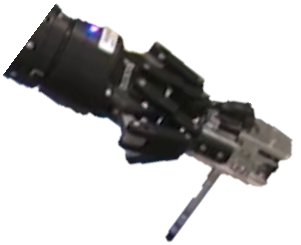}
    \includegraphics[height=100pt]{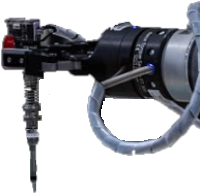}
    \caption{Hands and tools, from left to right: (a) Robotiq 2-Finger 85, (b) Precision gripper (custom), (c) Robotiq gripper holding a suction tool, (d) Robotiq gripper holding a screw tool}
    \label{fig:hands_and_tools}
\end{figure}

\subsection{Tools} 
\label{sec:tools}

We have designed a range of tools which are grasped by the general purpose grippers and then used by the robot arms, just as a human would pick up a tool and use it:
\begin{enumerate}
    \item 3 screw fastening tools (M3, M4, M6)
    \item 3 nut fastening tools (M4, M6, M10)
    \item 1 set screw tool
    \item 1 suction tool
\end{enumerate}

The grippers and some of the tools used are shown in Figure~\ref{fig:hands_and_tools}.
As this paper focuses on the kitting and assembly task, the set screw tool and nut fastening tools which were used in the taskboard task are explained in \cite{nie2019} in this same issue. 

\begin{figure}[!t]
    \centering
    \includegraphics[height=170pt]{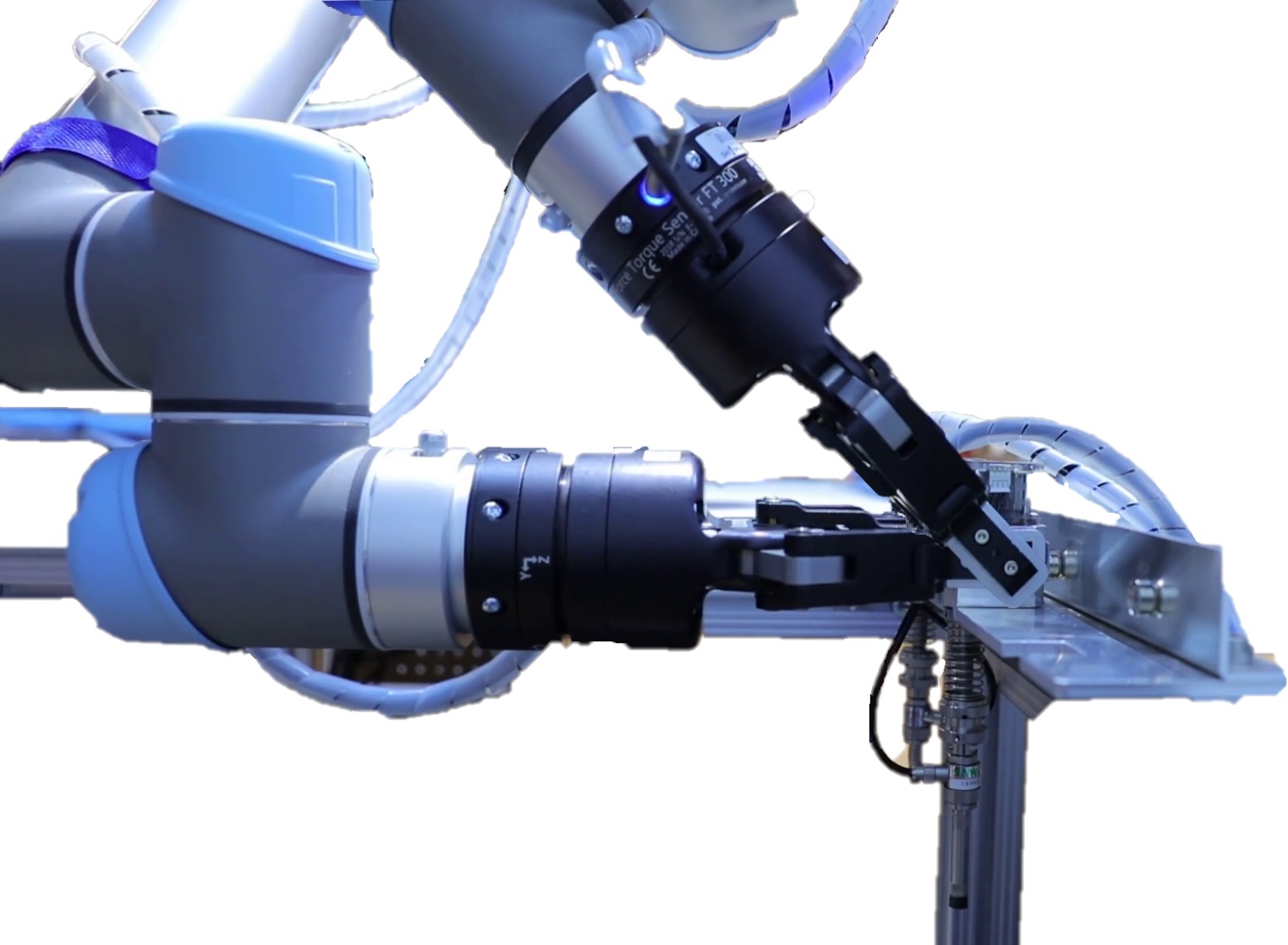}
    \includegraphics[height=170pt]{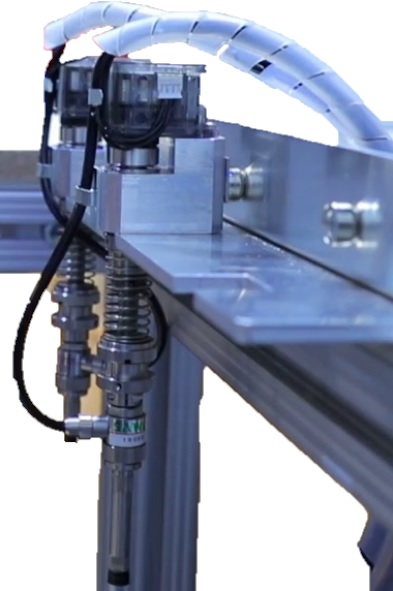}
    \caption{Right: The tool holder. Left: Different tool grasp orientations. The diagonal grasp orientation increases the usable range of robot C when holding the tool significantly.}
    \label{fig:tool_holders}
\end{figure}

All tools can be stored in the tool holder shown in Figure~\ref{fig:tool_holders}, which is located in between robots B and C, as shown in Figure~\ref{fig:robot-system-photo}.
In the tool holder, each tool rests on a surface which defines its vertical position, and is held in place with a magnet which defines its horizontal position.
This simple structure allows the tool to be placed and regrasped many times with high precision and repeatability.
The tools can be grasped by both robots B and C in different orientations, which significantly increases the useful range of motion of the smaller UR3 robot when the tool is held.
It also allowed alternative configurations to be chosen when the workspace around the assembly is crowded.

\subsubsection{Screw tools} 
\label{sec:screw-tools}

The screw tools are used to pick up, place and fasten screws.
The tip of each tool consists of a screw bit and a suction pad, which are spring-loaded and move independently from one another (both along and around the screw axis).
This allows the robot to apply pressure by moving the tool ``into'' the screw using position control instead of impedance control, using no complex robot hardware or control.

\begin{figure}[!t]
    \centering
    \includegraphics[width=.7\textwidth]{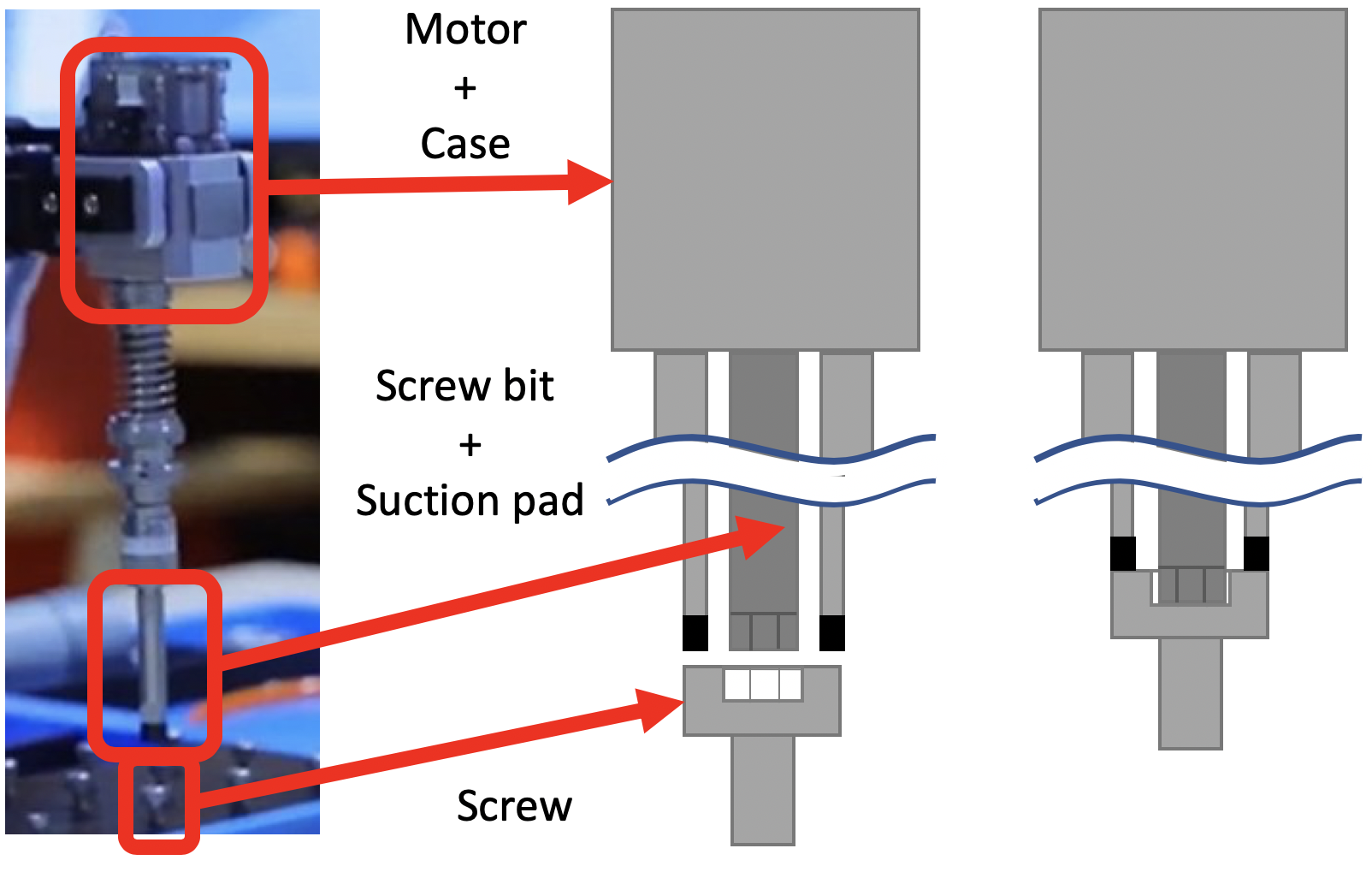}
    \caption{Both the screw bit and the suction cup at the tip of the screw tool retract independently when they are pushed onto the screw. On the right, the screw bit entered the screw's head and the suction cup has retracted to fit around the head}
    \label{fig:screw-tool-detail}
\end{figure}

To pick up a screw, the screw tool is moved ``into'' the screw while the motor is turned on and suction is applied, so that the bit enters the head of the screw and centers it inside the suction pad, as shown in Figure~\ref{fig:screw-tool-detail}. 
The successful connection of the screw is detected reliably by a change in air pressure.

To place the screw in the tray during the kitting task, pressurized air is used to eject the screw from the tip, since the remaining vacuum in the tube and tool would otherwise take too long to dissipate on its own.

To fasten the screw safely, the motor is equipped with a torque limiter.
The screw is considered to be successfully fastened if the motor has stopped turning and the pressure sensor indicates that the screw remained in position after the tool moved back from the screwing position.

The tools are connected by cable to a motor controller, and by tube to a vacuum tank and a pressurized air tank.
A U2D2 controller is used to connect to a Dynamixel motor, and a 3-way solenoid valve connected to the robot's controller box switches between suction, high-pressure and ambient air.

\subsubsection{Suction tool} 
\label{sec:suction-tool}

The suction tool is connected to the same vacuum tank as the screw tools, and is controlled using solenoid valves connected to the controller box of robot B.
Choosing a suction system involves the analysis of several parameters, such as the suction cup size and type, pressure difference and flow, as explained in~\cite{GarciaRicardez2017a} and its follow-up journal article\footnote{Accepted for publication, print date TBD}.
Because the surfaces of the objects were smooth and generally provided reliable seals, we chose a vacuum tank with a capacity of 39~l, and as most items to pick were relatively light, a suction cup of 9~mm diameter.

\subsubsection{Layout} 
\label{sec:layout}

While the robots and grippers did not change between tasks, different tools were selected in each task to fulfill the requirements, as shown in Table~\ref{tab:tool-setlist}.
Additionally, a 3D sensor was set up for the kitting task, and screw feeders for the taskboard and kitting task.

\begin{figure}[ht]
  \centering
  \includegraphics[width=160mm]{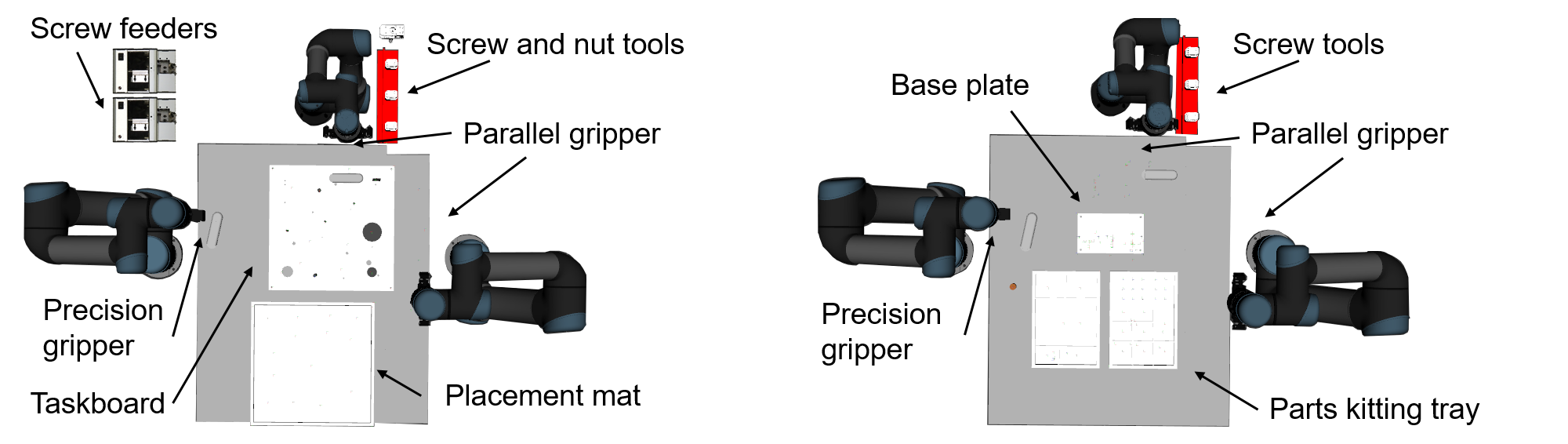}
  \caption{Top view of the scene layout in the taskboard (left) and assembly (right) tasks.}
  \label{fig:taskboard_assembly_layout}
\end{figure}

\begin{figure}[ht]
  \centering
  \includegraphics[width=120mm]{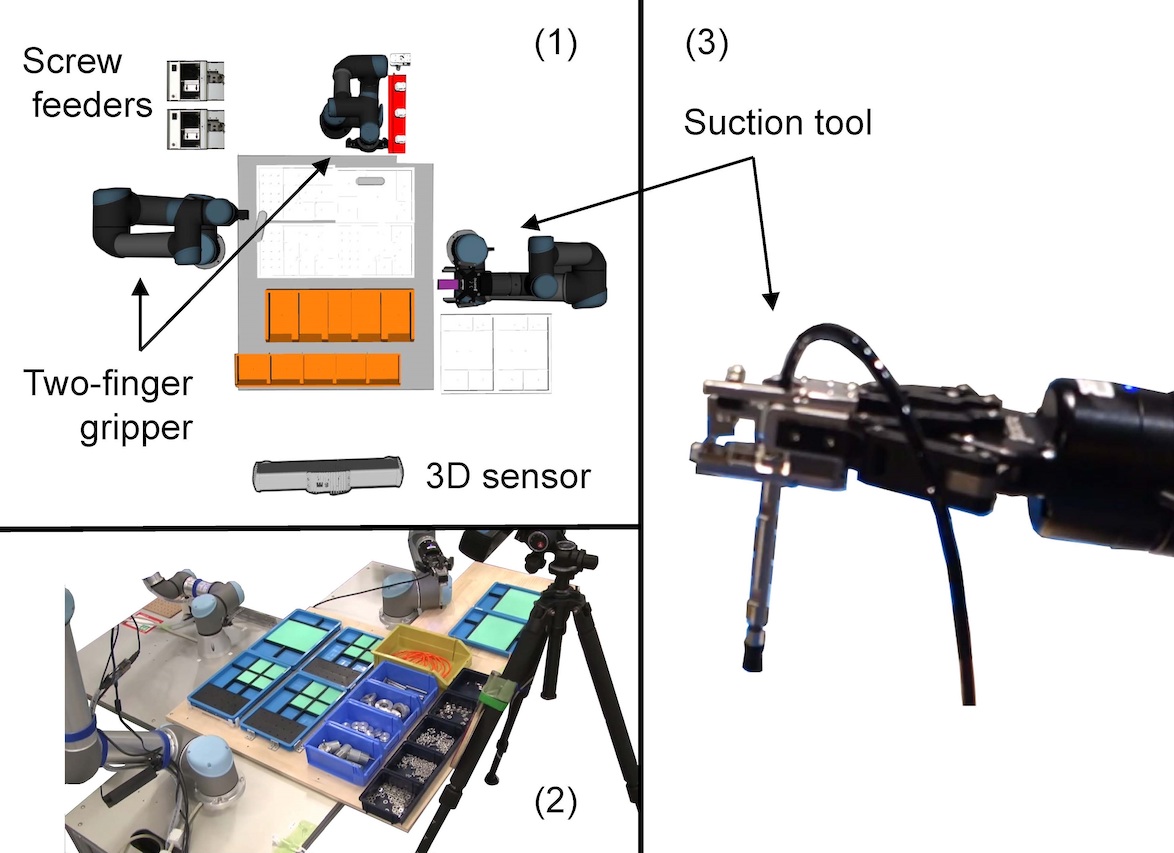}
  \caption{Layout of the robots, camera, bins and trays during the kitting task. All sets of trays were placed in the scene to avoid the need for resets. (1) Top view. (2) Overview. (3) Suction tool.}
  \label{fig:kitting_layout}
\end{figure}

The layouts in each task are shown in Figures~\ref{fig:taskboard_assembly_layout} and \ref{fig:kitting_layout}.

\begin{table}[ht]
\centering
\caption{Tool requirements for the different tasks}
\label{tab:tool-setlist}
\begin{tabular}{|l|l|l|l|l|l|l|l|}
\hline
Tool category & \multicolumn{4}{c|}{Screw tool} & \multicolumn{2}{l|}{Nut tool} & Suction tool      \\ \hline
Target parts & Set screw  & M3 & M4 & M6 & M6     & M8     & Other \\ \hline
Taskboard & $\checkmark$ & $\checkmark$ & $\checkmark$ & $\checkmark$ & $\checkmark$  & $\checkmark$ & $\times$  \\ \hline
Kitting & $\times$ & $\checkmark$ & $\checkmark$ & $\times$ & $\times$  & $\times$ & $\checkmark$  \\ \hline
Assembly & $\times$ & $\checkmark$ & $\checkmark$ & $\times$ & $\times$  & $\times$ & $\times$  \\ \hline
\end{tabular}
\end{table}

\subsection{Perception}

\subsubsection{Cameras}

Our system was equipped with two cameras.
One was a Photoneo PhoXi 3D Scanner Model-M fixed to the environment about 800~mm above the working surface. 
It can acquire 16 millions points per second as depth images with a $2064\times 1544$ resolution, and capture the 3D shape of both matte and reflective objects. 
We used the depth images thus obtained to detect graspable points on the parts in the kitting task, as detailed in Section~\ref{sec:kitting-task}.

The other camera was an Intel Realsense SR300\footnote{Creative blasterx Senz3d} attached to the wrist of robot A. 
Although it can record both depth and color images, we utilized only RGB images to determine whether the gripper successfully picked a part, as described in Section~\ref{subsubsec:checking_picks}. 
The camera was mounted on the penultimate link of the arm, which means its pose is not affected by the rotation of the last robot joint.

\subsubsection{Force sensing}
\label{sec:force-sensing}

Robots B and C were each equipped with a Robotiq FT300 force sensor, which was used to advance the robot until a physical contact occurred, and to move the bearing shaft screw forward during the insertion described in Section~\ref{sec:approach_idler_pulley}.
The robots were also capable of using the motor currents and FT300 sensor for impedance control during force-based insertion, to insert e.g. the bearing into the plate, and the peg into the bearing.
However, while we attempted to insert the peg into the bearing using a spiral search and impedance control, we were not able to arrive at a sufficiently reliable solution.
Positioning uncertainties, unstable grasps of the object (slippage during the operation), low control frequency\footnote{The UR robots could theoretically achieve a control frequency of up to 125 Hz, but were in practice limited by the execution and calculation load of the code.}, tight tolerances and other issues often caused the part to get stuck or the robot to apply too much force and thus enter a protective stop.
As these protective stops required user intervention to unlock, we did not perform force-based insertions of the bearing and peg during the competition.

\subsection{Software} 
\label{sec:system-software}

Our system was based on the Robot Operating System (ROS) and ran on C++ and Python.
The source code (excluding proprietary 3D mesh matching modules) is available online at https://github.com/o2as/ur-o2as.

\subsubsection{Containerized framework} 
\label{sec:containerized-framework}

As our team consisted of 4 subgroups in different locations, we needed to ensure that each group was working in the same software environment and on the same code.
To this end, we used a git-based workflow and a docker image that contained the virtualized environment that our robot solution is executed in.
This made it extremely easy for each member to reproduce the same development environment on their machines, without worrying about installing dependencies, software versions or network settings.
In turn, this allowed all members to develop code on their machines and test it in simulation, even while traveling or when the robots were unavailable.

The basic structure of this software solution is a git repository containing the ROS workspace, a set of scripts for installing and configuring the container, and a script to enter the container while mounting.
The container mounts the workspace folder, so that code changes are reflected both in the container and on the host machine, and the host's preferred development environment and editor can be used to write code.
The container itself is persistent, and only resets when changes are made to its installation instructions.

The solution on which our system is based is described in \cite{elhafi2018}.

\section{Approach} 
\label{sec:approach}

Our approach can be summarized as follows:
\begin{itemize}
    \item Avoid specialized jigs and keep the system flexible
    \item Expect, accept and deal with uncertainty (there will always be noise, calibration is never perfect)
    \item Use tools flexibly to extend the capabilities of the robot
\end{itemize}

In the following subsections we will describe in detail the approach followed for the assembly and kitting tasks.

\subsection{Assembly Task} 
\label{sec:assembly-task}
\begin{figure}[h]
    \centering
    \includegraphics[width=70mm]{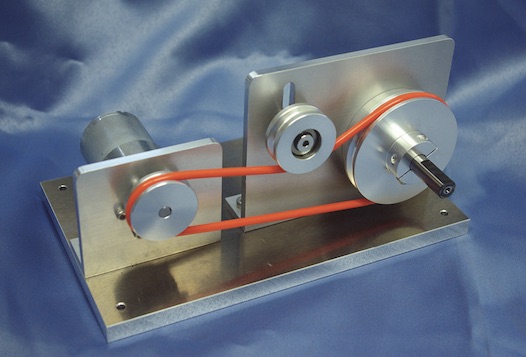}
    \caption{
        The drive unit to be assembled by autonomous robot systems {\cite{wrs2018_rulebook}. Courtesy of the WRS Industrial Robotics Competition Committee}}
    \label{fig:belt-drive-unit}
\end{figure}

In the assembly task, the robot system needs to autonomously assemble both the original belt drive unit, shown in Figure~\ref{fig:belt-drive-unit}, and another variant (``surprise belt drive unit'') for which the details were released only approximately 30 hours prior to the execution. 
This requires several problems to be considered. 
For one, the elementary skills required to complete the task and subtasks should be identified, so that they can be used as building blocks of the assembly instructions.
Secondly, the robotic system must be able to execute the basic skills and align objects. 
Lastly, the system must be easily reconfigurable, in order to assemble the surprise belt drive unit.

\begin{figure}[h]
    \centering
    \includegraphics[width=\textwidth]{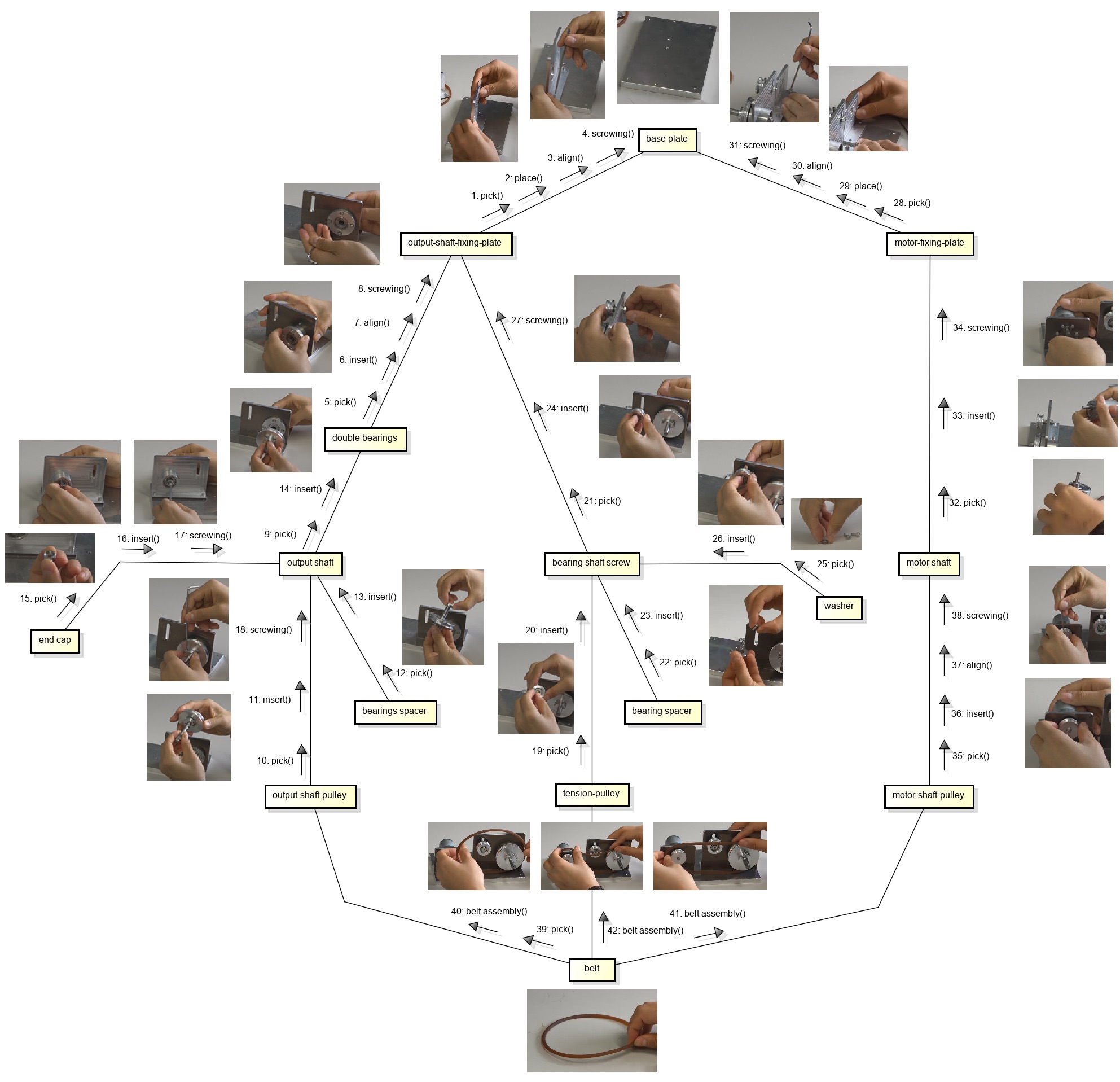}
    \caption{
        Assembly analysis from the video of a human assembling a belt-drive unit. Arrows indicate the order of operations to assemble a part, and the steps that need to be completed before a part can be mounted.
    }
    \label{fig:assembly_tree_analysis}
\end{figure}

The assembly task in the WRC 2018 comprises several basic skills, such as ``pick'', ``place'', ``insert'' and ``screw''.
We identified a list of such skills by analyzing the demonstration of a human performing the assembly of the belt-drive unit, as shown in Figure~\ref{fig:assembly_tree_analysis}. 
Based on this, we defined the structure of our software.
More details on the screwing task are given in Section~\ref{sec:screwing_task}.

Most of the objects (except the metal plates) are initially placed on two trays with partitions that separate each part. 
As the orientation and the position of each part could be freely chosen, as long as each part is placed in the correct compartment, we chose to place each object at the corner of each compartment at a specific orientation, so that their orientation and pose is known with relatively high accuracy. 
While this allowed us to forego object localization using vision, minor uncertainties (on the order of 1-2~mm) still resulted in small misalignments of the grasped parts. 
In section \ref{sec:approach_idler_pulley}, we explain how we dealt with such uncertainties in the case of the bearing shaft screw set.

In order to allow for a quick reconfiguration, we constructed an assembled Computer-Aided Design(CAD) model in the simulated environment based on which the relevant motions are defined.
This allows the poses to be adjusted automatically to the dimensions of new parts, making reconfiguration extremely simple and flexible. 
This is explained in more detail in the following section.

\subsubsection{Modeling and surprise parts approach} 
\label{sec:approach-surprise-parts}

As both the surprise belt drive unit and the known unit needed to be assembled during the same task, and as there would be very little time to teach any new positions to the robot, we aimed to make the transition between the known belt drive unit and the surprise parts as seamless as possible.
To achieve this, the following steps were taken:
\begin{itemize}
    \item define coordinate frames (``subframes'') systematically on all the parts in the assembly (e.g. ``part13\_screw\_tip'' or ``part4\_hole\_center''),
    \item create a module that defines a CAD-like assembly in the robot world, based on the parts' measurements and models, and
    \item parametrize all the operations in our assembly relative to the positions of the assembled parts - not in the world or robot base frame.
\end{itemize}

\begin{figure}[!htbp]
    \centering
    \includegraphics[width=\textwidth]{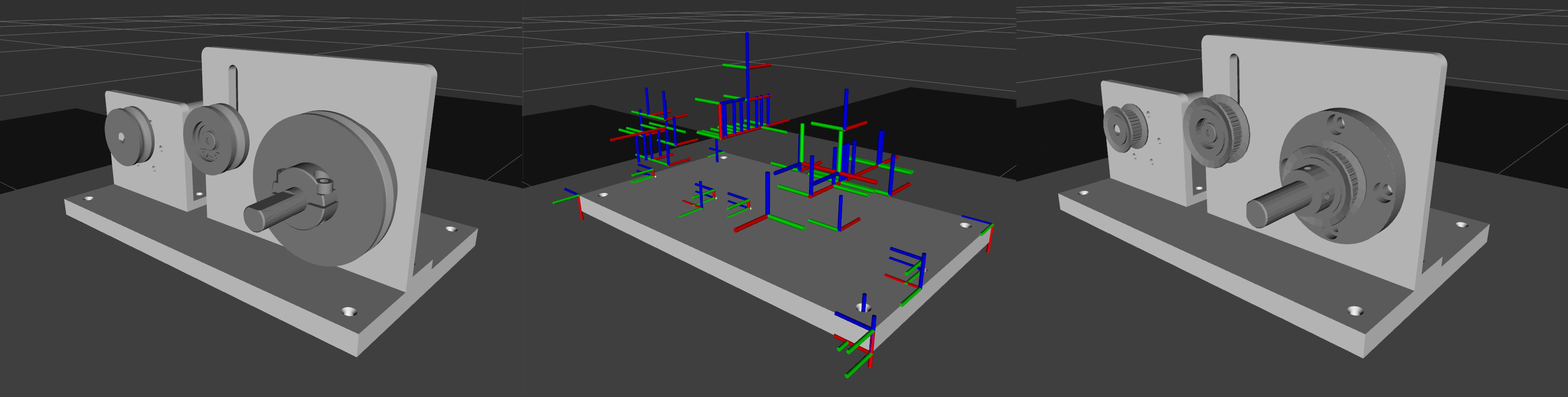}
    \caption{Left, right: The known and surprise parts as represented in the robot world. Middle: The coordinate frames that are defined on each object in the assembly, and which serve as the reference points for the assembly commands.
    }
    \label{fig:surprise-assembly-model}
\end{figure}

This way, the position that the assembly operation refers to is updated when a new CAD model or measurement is inserted into the environment.
This method was successfully implemented during the assembly task with surprise parts on the 4th day of the competition, as shown in Figure~\ref{fig:surprise-assembly-model}.
While only a part of the subtasks was executed by the robots, we confirmed the correctness of the new assembly in simulation.

The definition of the assembly frames was implemented through a list of connected frames in a text file, which our script uses to generate the tree of frames.
This simple interface can be extended with a module to import assembly files from step files or CAD software files.
A minor difficulty in this implementation was that transformation trees in ROS have to be strictly hierarchical.
We worked around this limitation by connecting objects only by their ``main'' frames, although the connection in the ``CAD assembly file'' (the file that the user writes) can be expressed simply using subframes.
An extension of this approach, which allows motions to be planned using subframes of objects that are grasped by the robot, has been merged into the open-source software MoveIt.
A detailed example of how this package can be used is presented in our code repository\footnote{https://github.com/o2as/ur-o2as}.

\subsubsection{Example: bearing shaft screw set assembly} 
\label{sec:approach_idler_pulley}

\begin{figure}[!htbp]
    \centering
    \includegraphics[width=\textwidth]{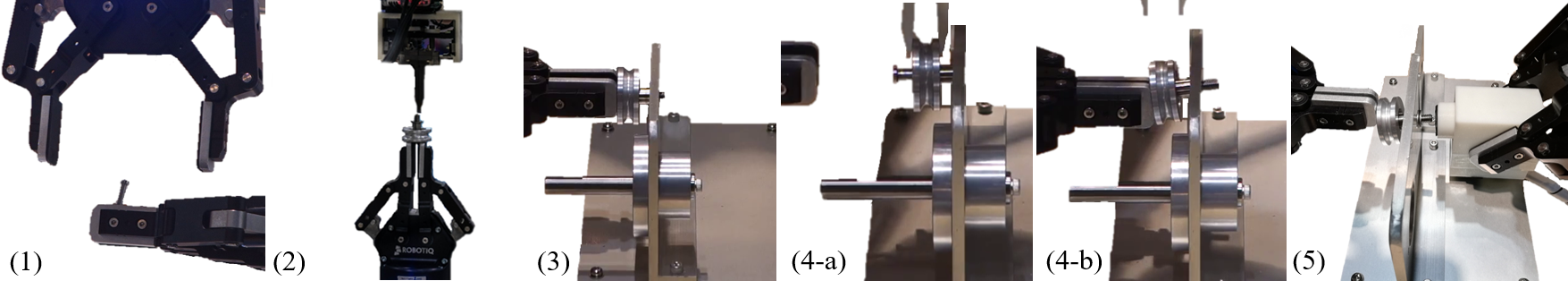}
    \caption{Snapshots of the assembly of the bearing shaft screw set.
    }
    \label{fig:assembly_bearing_shaft_screw}
\end{figure}

This section explains our approach to assembling the bearing shaft screw set (subtask E in the competition).
It can be considered as a series of peg-in-hole assemblies, where the same peg is inserted into multiple holes.
The bearing shaft screw (the peg) is inserted consecutively into the idler pulley, spacer, washer, metal plate, washer, and M4 nut.
We grasped all the objects to be assembled onto the peg from the inside with the precision gripper.
Our procedure for this subtask can be summarized like this:
\begin{enumerate}
    \item Pick the bearing shaft screw at the head and straighten/align it using the general purpose grippers (robots B and C)
    \item With the bearing shaft screw held upwards, pick the idler pulley, spacer and washer, and then release each part so it slides onto the bearing shaft screw
    \item Move the bearing shaft screw into the hole in the metal plate, stop when a force is encountered
    \item Use the precision gripper to hold the idler pulley. Open, move back and close the general purpose gripper, then push forward with it, so that the bearing shaft screw is pushed through
    \item Pick the washer with the precision gripper and drop it onto the threaded end of the screw, and pick the nut with the nut tool (held by robot C) and fasten it to the screw
\end{enumerate}
Figure~\ref{fig:assembly_bearing_shaft_screw} shows the snapshots of the assembly of the bearing shaft screw set explained above.

\begin{figure}[!htbp]
    \centering
    \includegraphics[width=\textwidth]{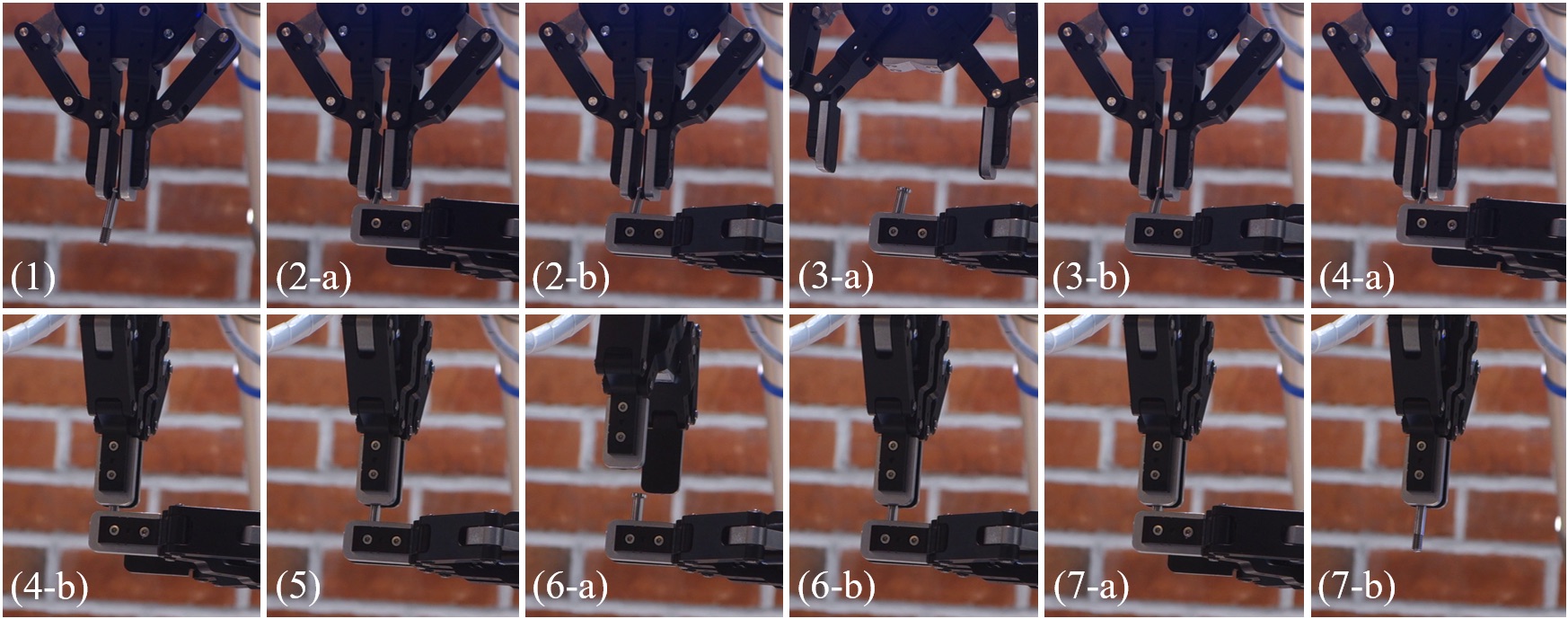}
    \caption{Snapshots of the bearing shaft screw being aligned using two general purpose grippers.
    }
    \label{fig:assembly_centering_bearing_shaft_screw}
\end{figure}

One point of difficulty is that the bearing shaft screw is often misaligned after the grasp.
The bearing shaft screw is initially placed upright in the tray, and the gripper grasps its head.
However, even small position inaccuracies cause the screw to be inclined, which shifts the position of the screw's tip considerably.
To alleviate this problem, we align the screw using a series of regrasps by robots B and C, as shown in Figure~\ref{fig:assembly_centering_bearing_shaft_screw}:

\begin{enumerate}
    \item Robot B (UR5 robot arm) picks up the bearing shaft screw at its head. 
    \item The bearing shaft screw is moved to a position which is reachable by robot C, and robot C grasps the shaft of the bearing shaft screw. This forces the screw to be upright within the grasp of robot C's parallel gripper.
    \item Robot B releases its grasp, and grasps the head of the bearing shaft screw again. 
    \item The gripper of robot C is opened, and the gripper of robot B is rotated by 90 degrees along the gripper's z axis.
    \item Robot C once again grasps the screw along the shaft. 
    \item Robot B opens and closes its gripper again, to grasp the screw at its head.
    \item Robot C opens its gripper and moves back to its initial position.
\end{enumerate}

After this procedure, the screw is aligned, unless the initial deviation was too significant.
This approach allowed us to reliably correct deviations of roughly up to 10 degrees.

Another difficulty when assembling the pulley and spacer on the bearing shaft screw is the fact that the screw rotates easily around the point of grasp.
The forces applied by the robot in impedance control would be large and easily rotate the screw.
Instead, we chose to use gravity and the passive compliance of the precision gripper to slide parts onto the screw: we pick the parts from the inside with the precision gripper, position them above the screw, close the precision gripper to release the part, and initiate a spiral motion that ensures that the item slides onto the screw shaft.
This applies only small forces, which do not affect the screw's position, and assembled the parts with high repeatability.

\subsection{Kitting Task} 
\label{sec:kitting-task}

\subsubsection{Kitting Task Procedure}
\begin{figure}[ht]
  \centering
  \includegraphics[width=80mm]{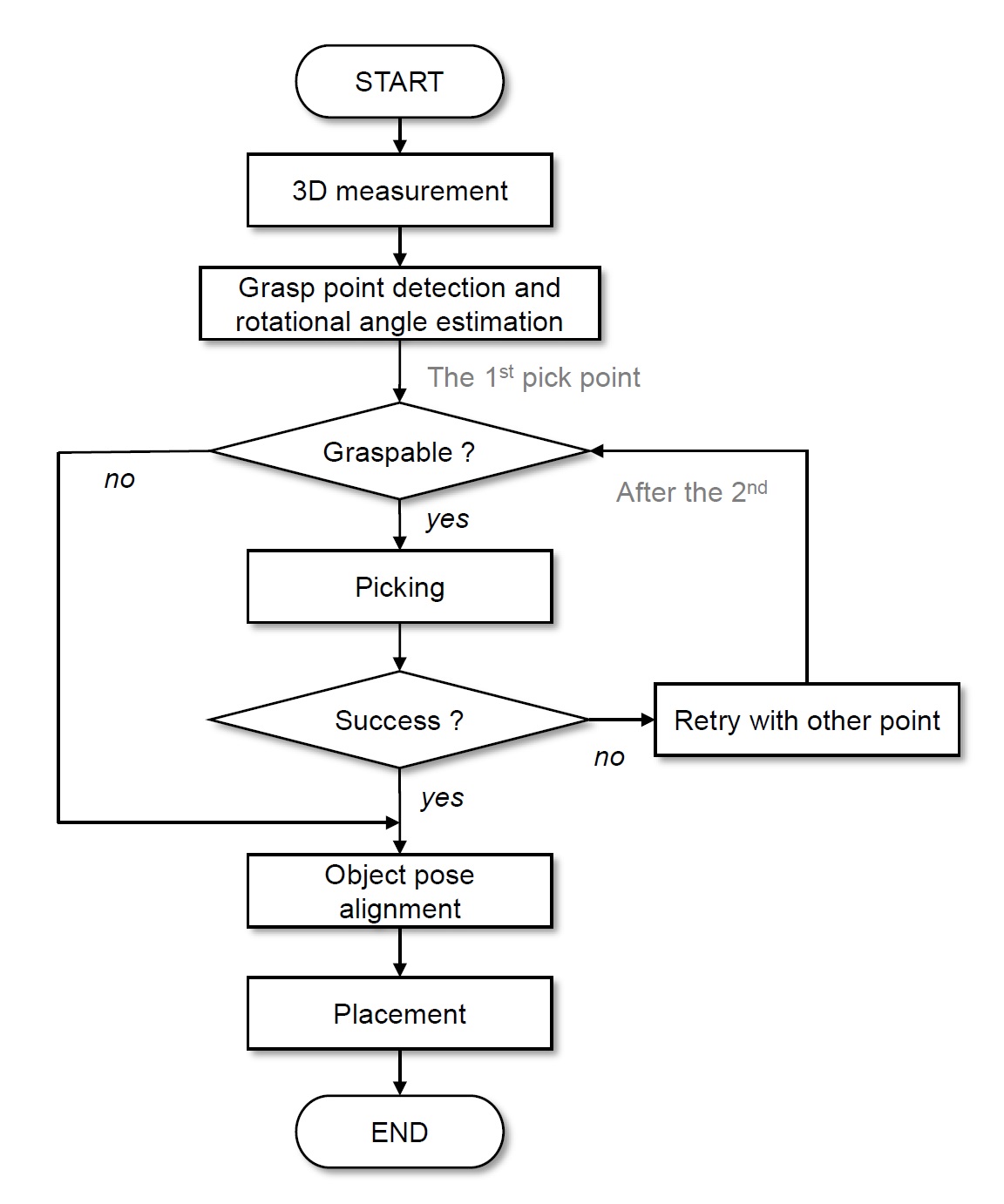}
  \caption{Flowchart of the kitting task.}
  \label{fig:kitting_flowchart}
\end{figure}

The kitting task is composed of bin-picking tasks and pick-and-place tasks for target parts that are small and shiny, e.g. screws, nuts, etc. 
The full set of rules is detailed in \cite{wrs2018_rulebook}. 
Our robotic system and the flowchart for the kitting task are shown in Figure~\ref{fig:kitting_layout} and Figure~\ref{fig:kitting_flowchart} respectively. 
First, we acquire a 3D image of the objects in the bins and determine a valid, graspable position and a corresponding gripper pose, at which the robot then picks the target part. 
During the kitting task, robot arms A and B are used to pick the items, while robot C uses the screw tool to place screws in the tray.
Screws are picked using the two-finger precision gripper, placed in one of the screw feeders, and then picked and placed using a screw tool.
Robot B carries the suction tool described in Section~\ref{sec:suction-tool}, robot A is equipped with the two-finger precision gripper (see Section~\ref{sec:system-hardware-robots-and-grippers}), and robot C uses the screw tools. 
We manually defined which gripper is used to pick each part. 

\begin{figure}[ht]
  \centering
  \includegraphics[width=40mm]{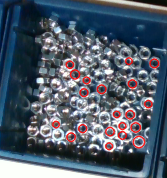}
  \caption{Result of searching for the graspable hole in M4 nuts, based on 2D contour detection method.}
  \label{fig:kitting_blob}
\end{figure}

The main challenge in the kitting task is to detect graspable points from cluttered scenes. 
Existing pose estimation algorithms which use 3D CAD models of parts~\cite{Besl1992,Mliu2010,Drost2010,Choi2012} could solve this problem, but these methods are difficult to apply to small and shiny parts, such as screws, nuts, etc. 
Also, current pose estimation methods are strongly affected by sensor characteristics such as resolution and noise, ambient light settings, object surface states and sizes, and more.
In Figure~\ref{fig:kitting_blob} we show the result of a 2D image processing method based on contour detection to detect the graspable hole of parts like nuts and washers, but obtained mixed results due to lighting and reflections affecting the detected contour, and reflections and edges affecting the quality of the measured point cloud.
Therefore, we selected methods that directly find graspable points without pose estimation or contour detection.

Grasping point detection algorithms~\cite{Mano2019,Matsumura2019,Domae2014} are robust methods to find graspable points from cluttered scenes. 
Fast Graspability Evaluation (FGE)~\cite{Domae2014} is a practical-use method that finds graspable points by convolving the 2D gripper model and the parts surfaces on a depth map. 
This method is applicable to both suction and two-finger grippers, and suitable for pick-and-place problems such as the kitting task.
 
However, FGE is limited in the sense that the method assumes that the gripper picks the target part from the same direction that the camera views the scene.
The target pose of the gripper is described using 4 degrees of freedom (x, y, z, z-axis rotation), therefore this method cannot be applied to objects that need to be grasped at an inclination relative to the camera plane. 
In our proposed system, robot arms are holding suction tools during the kitting task. In some cases, such 4 degrees of freedom approaches may be hard to use while holding tools.

\begin{figure}[ht]
  \centering
  \includegraphics[width=80mm]{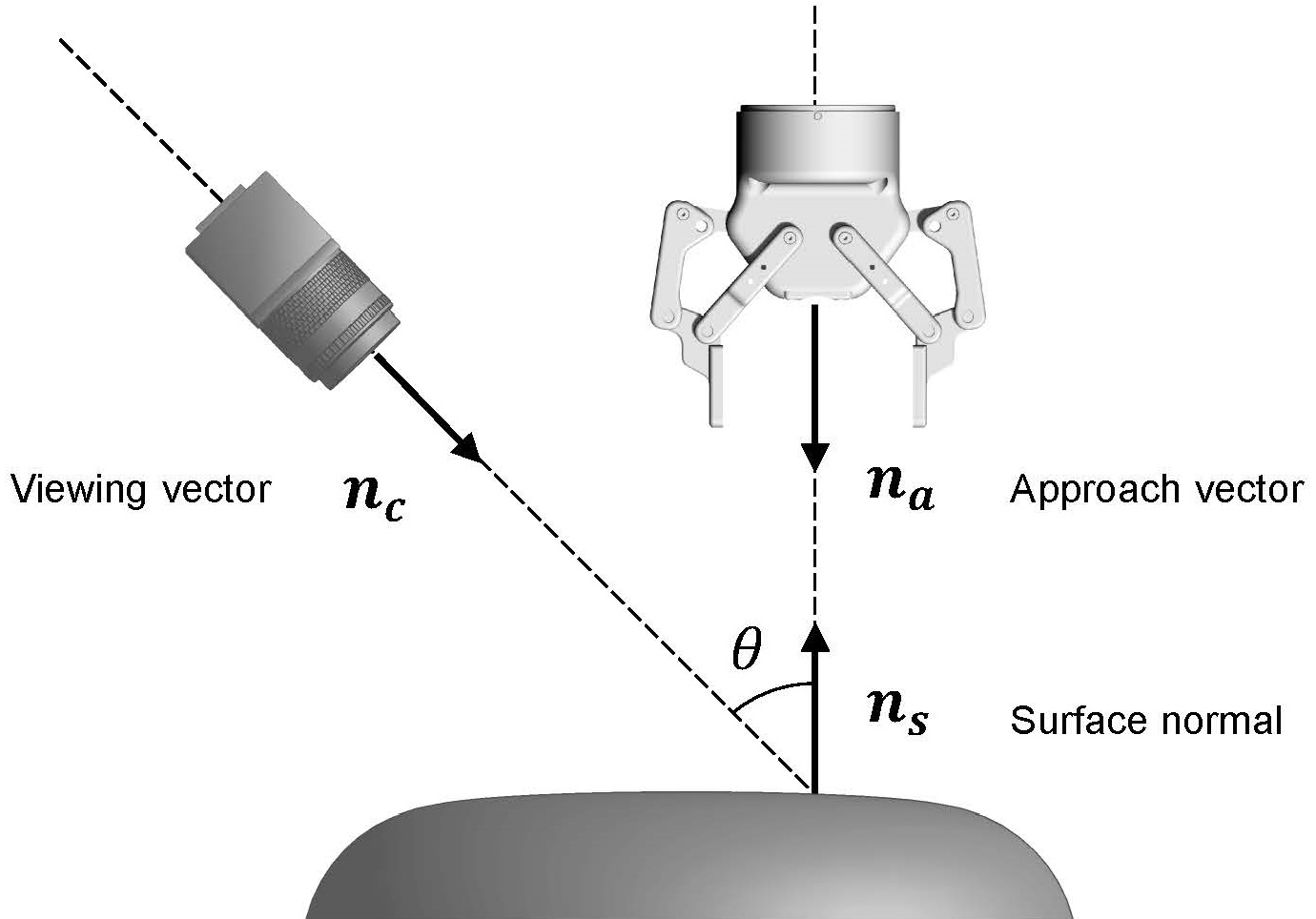}
  \caption{Correspondence between the viewing vector, the surface normal vector, and the approach vector.}
  \label{fig:kitting_approach}
\end{figure}

Therefore, we propose a new graspability metric ${\bm G}_{6DoF}$ using a single depth map as below, to apply these algorithms based on 4 degrees of freedom to pick items by 6 degrees of freedom motions. 

\begin{equation}
  {\bm G}_{6\textrm{DoF}}(x,y) = \omega_{\theta} {\bm G}_{4\textrm{DoF}}(x,y) \, ,
  \label{eq.6dof-graspability}
\end{equation}

\begin{equation}
  \omega_{\theta} = 
  \begin{cases}
      0 & \textrm{if} \hspace{2mm} \| \theta \| \geq T_{\theta} \\
      1-\frac{1}{T_{\theta}} \arccos{\left( \frac{-\bm{n}_c \cdot \bm{n}_s}{\| \bm{n}_c \| \| \bm{n}_s \| } \right)} & \textrm{otherwise}
  \end{cases}
      \, ,
  \label{eq.omega-detail}
\end{equation}

\noindent where $\bm{G}_{4DoF}$ denotes a graspability metric calculated by the 4-degrees-of-freedom-methods like FGE, and $\theta$ denotes the angle between the two vectors: $\bm{n}_c$ (a viewing vector by the camera), and $\bm{n}_s$ (a surface normal vector of the grasping point on the depth map). $T_{\theta}$ is a threshold angle. 
If $ \| \theta \| $ is too large, the quality of the depth image in the region of interest becomes low,  due to the optical principles of the cameras. 
Accordingly, in our experiment $T_{\theta}$ is set manually to $\pm$~25 degrees.
In addition, we define the approach vector of the gripper for picking as ${\bm n}_{a} = -{\bm n}_{s}$.
The correspondence between these vectors is shown in Figure~\ref{fig:kitting_approach}. 
We calculate the 6 degree-of-freedom pose of the approach gripper ${\bm P}_{g}$ for picking by finding the maximum ${\bm G}_{6DoF}$ from candidates in the depth map.

\begin{equation}
  {\bm P}_g = \left( \bar{x},\bar{y},\bar{z},\bar{n}_{gx},\bar{n}_{gy},\bar{n}_{gz} \right) \, ,
  \label{eq.poseby6dofgraspability}
\end{equation}

\noindent where $\bar{x},\bar{y},$ and $\bar{z}$ denotes a position which has a maximum peak of $\bm{G}_{6DoF}$, and $\bar{n}_{gx},\bar{n}_{gy},$ and $\bar{n}_{gz}$ denotes the approach vector of the gripper for picking which has a maximum peak of $\bm{G}_{6DoF}$.

\begin{figure}
    \centering
\includegraphics[width=130mm]{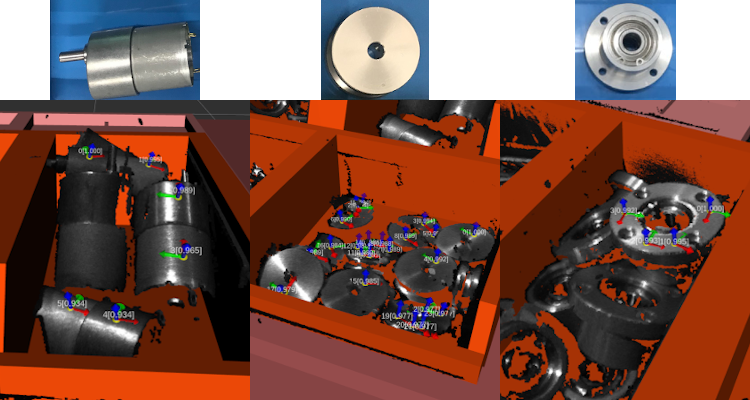}
    \caption{Grasping pose recognition results by using the proposed method and a suction tool model.}
    \label{fig:recognition-results}
\end{figure}

In Figure~\ref{fig:recognition-results}, we show the grasping pose recognition results on three parts (a gear motor, a pulley and a bearing). 
Grasping points for the suction tool are detected on each part.

\subsubsection{Hand-eye Calibration}

Hand-eye calibration, i.e. establishing the geometrical relationship between a robot arm and a 3D vision sensor, is an essential task that needs to be completed in advance of executing the kitting task. 
To this end, we used a calibration technique implemented in the {\em easy\_handeye} software package\footnote{https://github.com/IFL-CAMP/easy\_handeye}, in which an AR marker held by the gripper is presented to the camera at a number of different (we choose 30) positions and orientations.
The camera pose is then estimated with respect to the robot base frame from the pairs of marker poses observed by the camera and the corresponding end effector poses obtained from the robot kinematics. 
In order to improve numerical stability, we replaced the classical hand-eye calibration algorithm~\cite{Tsai1989} adopted by the original package with our own implementation which is formulated with quaternions instead of Euler angles to represent the rotational part of the end effector and camera poses.

\subsubsection{Checking for successful picks} 
\label{subsubsec:checking_picks}
In the kitting task, in order to increase the robustness of the system, we check if the pick was successful after every attempt to pick a part. 
We implemented three different checks: one for the suction tool, one for the precision gripper, and one for the general purpose gripper.

To check for picking success of the suction tool, we measure the air pressure inside the tube, as the pressure is significantly lower when an object is successfully attached to the suction cup and air cannot flow in.
If the measured value is below a certain threshold (which we set to -55 kPa), we assume that the item was successfully picked.

\begin{figure}[!htbp]
    \centering
    \includegraphics[width=\textwidth]{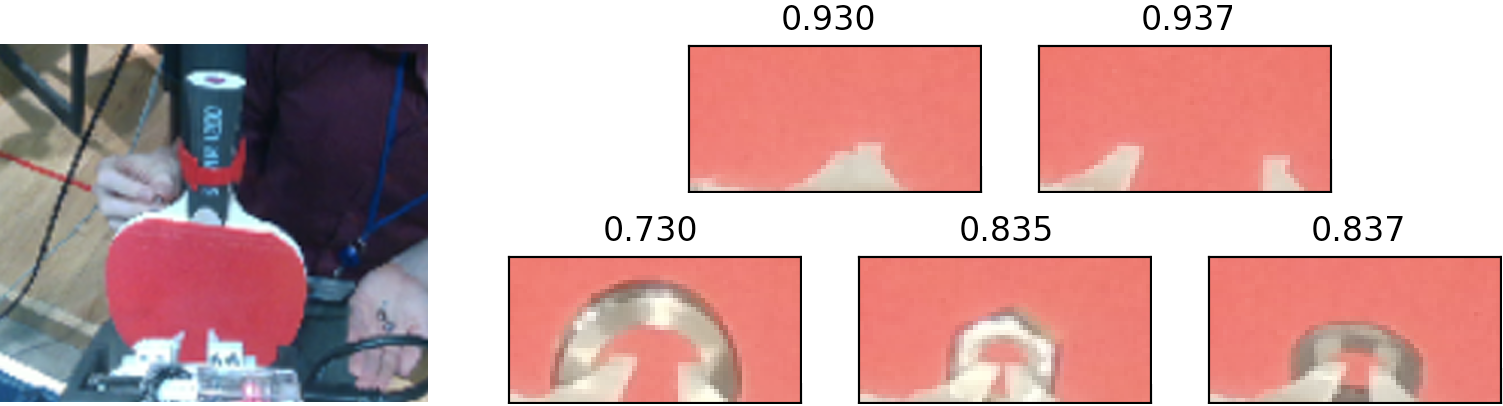}
    \caption{Left: The robot holds the gripper tip in front of a table tennis racket to check for grasp success using the average RGB value. Right: $R_{mean}$ values for different items (bottom) and empty grasps (upper).
    }
    \label{fig:table-tennis-racket-color-view}
\end{figure}

The grasp success for the general purpose gripper can be determined by checking the grasp width.
If the gripper is not fully closed after lifting it out of the bin and attempting to close it again, an item has been grasped.
However, in our precision gripper's prototype, the gripper opening width could not be determined with sufficient precision, because the materials were too flexible and the 3D printing tolerances too large. 
We thus chose to use an RGB-D camera mounted on the robot's wrist which observes the gripper, and a simple color matching procedure to visually determine the successful pick of an item. 
After each pick, the robot moved in front of a red surface as shown in Figure~\ref{fig:table-tennis-racket-color-view}, which facilitates the procedure to determine if an item is picked:
\begin{enumerate}
    \item Crop the image to a region of interest (ROI)
    \item Calculate $R_{mean}$ (average ``redness'' value) for the pixels in the ROI
    \item If $R_{mean} > R_{threshold}$, an item was picked.
\end{enumerate}

While this procedure is simple and easy to configure for known items and lighting conditions, it would be unnecessary if the gripper was manufactured with precision and supplied an accurate measurement of its opening width.

\section{Experiments} 
\label{sec:experiments}

\begin{table}[]
\centering
\caption{Team O2AS competition results}
\label{tab:score}
\begin{tabular}{|l|r|}
\hline
                                  & \multicolumn{1}{l|}{Score} \\ \hline
Taskboard task                    & 36                         \\ \hline
Kitting task                      & 13                         \\ \hline
Assembly task                     & 25                         \\ \hline
Assembly task with surprise parts & 34                         \\ \hline
Total                             & 108                        \\ \hline
\end{tabular}
\end{table}

In this section we show the experimental results obtained at the competition. 
In the kitting and assembly tasks we performed two trials per task (each trial lasting 30 minutes including setup time), while for the assembly with surprise parts we performed only one trial of 60 minutes (including setup time).
The total score achieved at the competition is shown in Table~\ref{tab:score}.

\subsection{Screwing Task} 
\label{sec:screwing_task}

To evaluate the performance of the screw tools under ``real'' conditions, we counted and timed the screw attempts during the actual WRC assembly trials.
We assume that this is the closest we can get to the conditions and constraints that it was designed to meet, and thus gives the most accurate representation of our system's performance.

As we did not assemble the motor and bearing during the competition, this count only includes attempts to screw the plates to the base plate, which use the M4 tool.
While this means that the M3 tool was not part of this performance evaluation, we have noticed no difference in reliability or performance during our pre-competition testing for the kitting task, where M3 screws had to be picked and placed into the trays.

The success rate for picking up and fastening screws during the assembly task trials was 24 out of 27.
The failures were caused by:
\begin{itemize}
    \item drops because of low air pressure and an insufficiently inserted drill bit. The vacuum pressure in the tank could vary and was slightly less reliable at the low end,
    \item the screw not being detected because of insufficient vacuum pressure, and
    \item large positioning errors of the tool.
\end{itemize}

\begin{figure}
    \centering
    \includegraphics[width=130mm]{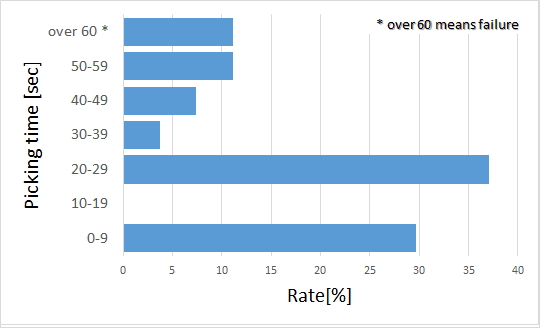}
    \caption{Time taken to pick an M4 screw from the tray during the competition.}
    \label{fig:pickingtime}
\end{figure}

As shown in Figure~\ref{fig:pickingtime}, the picking times show large amounts of scattering because of the spiral motion (``pecking'') used to pick up the screw, which increases the picking time drastically with a growing distance from the screw head.
We assume that performing the spiral motion without pecking would have been quicker, but decided against it, as it would risk shifting the tray, or the tool inside the gripper.

\subsection{Parts Kitting}

\begin{figure}[ht]
  \centering
  \includegraphics[width=160mm]{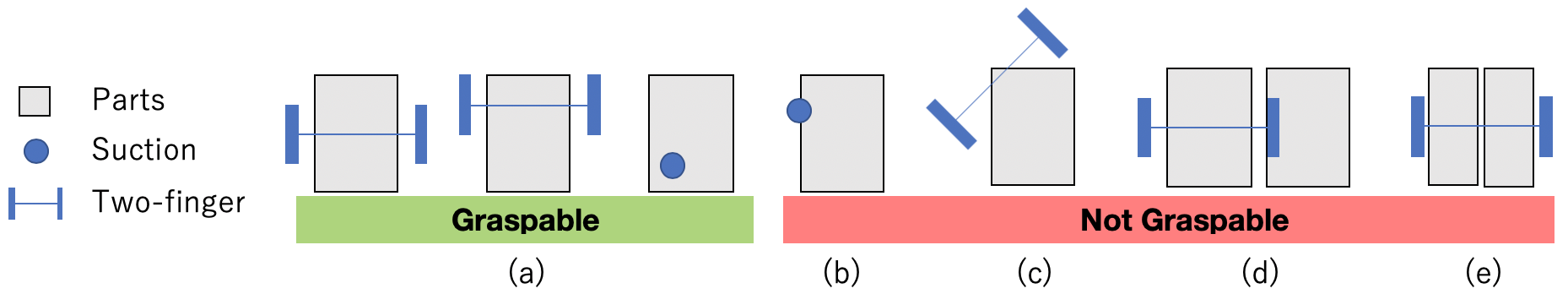}
  \caption{Definition of graspable points and no graspable points. (a) examples of graspable points, (b-e) examples of non-graspable points. (b) point at an edge of the part, (c) escaping part during closure of the gripper, (d) collision, gripper's finger is over other parts (or surroundings), (e)  multiple parts picking.}
  \label{fig:definitionofgraspablepoints}
\end{figure}

In this subsection, we evaluate our proposed algorithm performance for grasping point detection by using depth maps with multiple parts. We define the rate of valid grasp candidates as, 

\begin{equation}
  \left( 1-\frac{\text{The total number of non-graspable points in the depth maps}}{\text{The total number of grasping points extracted in the depth maps}} \right) \times   100
  \, .
  \label{eq.graspable-recognition-rate}
\end{equation}

First, the grasping points are detected automatically from depth maps using the proposed algorithm, as described in Section~\ref{sec:kitting-task}.
Attempting to evaluate a grasp point physically with the robot would not only disturb the arrangement of items, so that only one grasp candidate can be evaluated in each view, but it would also be unclear if the pick failure was due to an unrelated circumstance, such as a shifted calibration, insufficient air pressure, a broken suction cup, positioning noise etc.
To avoid letting these factors influence our evaluation, we choose to manually evaluate the points proposed by the algorithm to determine if they are graspable or not.
We define the graspable and non-graspable points as shown in Figure~\ref{fig:definitionofgraspablepoints}.
Graspable points are defined as explained in~\cite{Domae2014}. 
Non-graspable points, shown in Figure~\ref{fig:definitionofgraspablepoints} (b)-(e), are:
\begin{itemize}
    \item those that are on a part's edge or not on the part's surface (b),
    \item ones that are in a location where the part would slip from the grasp after closing it (c),
    \item points that overlap other parts or the surrounding environment (d), 
    \item and points that enclose multiple parts (e).
\end{itemize}

\begin{figure}
    \centering
  \includegraphics[width=130mm]{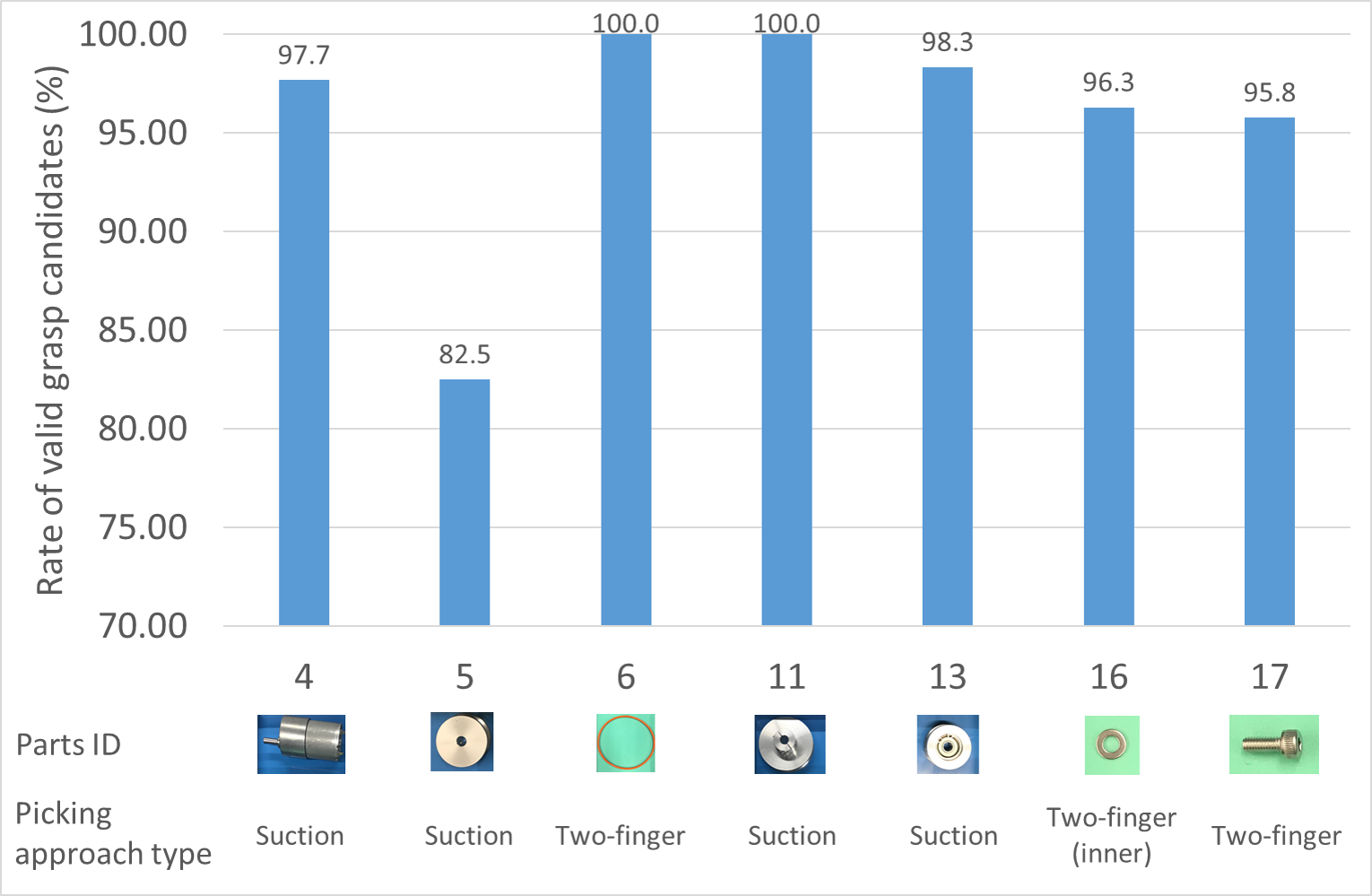}
    \caption{Rate of valid grasp candidates for 7 different parts when using the proposed grasping point detection algorithm.}
    \label{fig:recognitionRate}
\end{figure}

The total number of evaluated grasping points in the experiment is 302. 
The experimental results are shown in Figure~\ref{fig:recognitionRate}. 

We use three types of gripper approaches: suction tool, two-finger, and two-finger (inner). 
The ``inner'' type means that the two-finger gripper is closed during the approach and then opened to grasp the part from the inside. 
We use three gripper models to detect grasping points from depth maps.

In spite of the difference between the viewing vector and the approach vector in this system, the recognition rate of grasping points in the images is high, as is the recognition rate for each picking approach type. 
The rate of part ID 5 (small pulley/idler bearing) is lower than the other parts' rates, because many detected points are on the edge or outside of the part. 
While the recognition rate depends on the parts' shape, the success rate is still over $80\%$. 
Thus, we conclude that the algorithm is capable of detecting grasping points in our proposed system configuration.

\section{Discussion} 
\label{sec:discussion}

Our experimental results demonstrate the validity of our proposed system, but as the system was built entirely from scratch over the course of 6 months, it was not mature and there was little time to test and program it.
Thus, we consider it a proof of concept and believe that the potential is higher than its competition score suggests.

We consider the strongest points of our system to be the flexibility of our software, our approach to using tools, and our integration of physical compliance.
Modeling the full assembly (excluding screws) in kinematic simulation and basing all of our assembly operations on these models allowed us to incorporate new parts quickly and systematically.
Furthermore, our containerized development environment allowed us to synchronize our workstations and streamline the development.
The screwdriver and suction tools allowed us to extend the capabilities of our robots without expensive tool changers, and made the system significantly more flexible.

The compliance helped us deal with uncertainty, and allowed us to implement very simple methods to center, position and pick up parts.
An example that demonstrates the advantages of this approach is the handover of the end cap, in which one gripper holds the end cap, and the closed precision gripper is moved in a spiral on it until the tip enters the end cap bore, even if the end cap's position is offset by several millimeters.

As noted in Section~\ref{sec:force-sensing}, we did not use force-based insertion during the competition to avoid protective stops and manual resets.
This is at least partly due to the robots' maximum control frequency of 125 Hz, and limited our system to certain subtasks, which is a flaw that we will tackle in the future.
We recognize that comprehensive force sensing can make the system significantly more robust and is a very promising, if not necessary component of an assembly system.
We note that physical compliance and impedance control can be used together to extend the range of a system's capabilities.

In terms of lessons learned, we have found that investing into debugging and visualization tools early helps speed up the development process by allowing easier introspection, deeper understanding of the system, and quicker fixing of bugs.
Especially parameters that are relevant to the robot system's decision-making and which can be visualized in camera images or 3D simulation can be insightful, such as target grasping points and motion targets.

Furthermore, we found that the development, especially the initial setup and creation of the system and framework, required a lot of work and time.
To groups implementing new solutions, we would suggest to use standard methods, existing work and lightweight prototyping using Python as much as possible.
To this end, we released our code open-source along with this paper as a point of reference: https://github.com/o2as/ur-o2as

\subsection{Kitting task}

Unexpectedly, our biggest difficulty in the actual kitting task was camera calibration, both for the scene-mounted 3D camera used for the grasp point detection and the robot-mounted RGB-D camera used for blob detection.
One possible explanation is that our 3D camera was mounted on a free-standing tripod instead of a fixed aluminum frame, which meant that the calibration could be lost when it was moved or bumped slightly.
The calibration procedure being time consuming to repeat was another issue during the competition.
In the future, we would make sure to implement a reliable calibration procedure that requires less than 5 minutes to finish and (possibly more importantly) attach the camera more rigidly relative to the robots.

Our kitting approach did not determine the orientation of the picked object either before or after the grasp.
As mentioned in the previous section, the grasping point detection algorithm depends on the shape and distribution of the parts inside the bin, but does not retrieve the orientation of the picked object.
As the orientation was required for the reliable placement of some items in the tray, our approach was incomplete in this regard.

The background color-based method we used to confirm the grasp success for small items (screws, washers, nuts, etc.) with the precision gripper proved sensitive to lighting conditions, so we would either replace it with a more robust method or remove the necessity for it by investing into a prototype that can measure the gripper opening width reliably.

\subsection{Assembly task}

Overall, we found that the amount of time required to set up the assembly instructions was very large.
One reason was that small calibration or positioning errors often caused operations to fail without the robot system perceiving this failure.
It appeared highly desirable to us if the robot system had an understanding of the operations to be performed with a part, how to check that each operation has succeeded, and possibly engage in a recovery strategy (e.g. look for a fallen part).
As of yet, there is no standard framework for this sort of procedure.
It is an open problem for the robotics community to agree on the standards and requirements for such a framework.
As a mid- to long-term research goal, an abstract task language seems to be a desirable tool to tackle this problem.

We found that the procedure to align parts using only the grippers without jigs described in Section~\ref{sec:approach_idler_pulley} was reliable and flexible, but required a considerable amount of time.
Concave fingertips have been proposed as a way to avoid the need for this, but we suspect that such fingertips would either be too slightly curved to have an effect, or be so specialized to certain parts that the shape may be an obstacle when grasping other parts.
It is true that both fingertips and jigs that are shaped for specific parts speed up the alignment process, but we did not include them in our workflow as they do not seem feasible for large part varieties.
In our next design, we would consider the use of either an actively controlled centering or alignment aid, or versatile passive jigs, e.g. calibrated surfaces and slopes that can be used for multiple parts.

Another relevant issue was robot and tool calibration, which is a fundamental procedure that is important to verify, additional to the camera calibration mentioned above.
It is easy to overlook mistakes in the calibration of tools and robot frames if the tests do not include rotated poses, if parts of the tools can shift or deform, or if tools are not reliably stored and retrieved in the same position.
Calibration errors due to either of these issues could lead to offsets of 2-4~mm at the tool tip, which was enough to cause problem when picking up small parts.
We suggest to use a maximum of care and attention to detail when setting up the calibration procedures.

\section{Conclusions}
\label{sec:conclusion}

We presented a robotic system that we have developed for the World Robot Summit Assembly Challenge 2018, which is capable of performing all the tasks in the competition, during which it obtained 4th place as well as the SICE (Society of Instrument and Control Engineers) Special Award.
The strongest points in our approach are the full modeling of the assembly system, the tools, and our software framework, which is available open source.
Our main contributions are the jigless nature of the system, the parametrization of the assembly, screw tools which can be grasped by parallel robotic grippers, an efficient grasp point detection algorithm and the source code running our solution.
In future work, we will focus on simple and reliable calibration, an automatic assembly approach, the ability to change part specifications on the fly, and extend the tools available to the robot.

\section*{Acknowledgment}
This paper includes work of a project commissioned by the New Energy and Industrial Technology Development Organization (NEDO), Japan. 

We are grateful to Yuma Hijioka, Xinyi Zhang, Yasushi Kowa, and Michiaki Hirayama for their contribution to this project (honoring titles omitted and in random order).

\bibliography{ma}

\end{document}